\documentclass[10pt,a4paper,twocolumn]{article}
\usepackage{hyperref}
\usepackage{times}
\usepackage{epsfig}
\usepackage{graphicx}
\usepackage{amsmath}
\usepackage{amssymb}
\usepackage{microtype}
\usepackage{graphicx}
\usepackage{amssymb, amsmath}
\usepackage{amsmath,amssymb,amsthm}
\usepackage{xfrac}

\usepackage{algorithm}
\usepackage{algorithmic}
\usepackage{float,mathtools}

\theoremstyle{definition}

\usepackage{microtype}
\usepackage{graphicx}
\usepackage{subcaption}
\usepackage{booktabs} 

\usepackage{enumitem}

\usepackage{amsmath,amssymb,amsthm}
\usepackage{booktabs} 

\usepackage{multirow}
\usepackage{listings}

\newcommand{\R}{\mathbb{R}}

\usepackage{hyperref}

\begin{document}

\title{PruneNet: Channel Pruning via Global Importance}

\author{Ashish Khetan\\
Amazon AWS\\
\texttt{khetan@amazon.com}
\and
Zohar Karnin\\
Amazon AWS\\
\texttt{zkarnin@amazon.com}
}
\date{}

\maketitle

\begin{abstract}
Channel pruning is one of the predominant approaches for accelerating deep neural networks. Most existing pruning methods either train from scratch with a sparsity inducing term such as group lasso, or prune redundant channels in a pretrained network and then fine tune the network. Both strategies suffer from some limitations: the use of group lasso is computationally expensive, difficult to converge and often suffers from worse behavior due to the regularization bias. The methods that start with a pretrained network either prune channels uniformly across the layers or prune channels based on the basic statistics of the network parameters. These approaches either ignore the fact that some CNN layers are more redundant than others or fail to adequately identify the level of redundancy in different layers. In this work, we investigate a simple-yet-effective method for pruning channels based on a computationally light-weight yet effective data driven optimization step that discovers the necessary width per layer. Experiments conducted on ILSVRC-$12$ confirm effectiveness of our approach. With non-uniform pruning across the layers on ResNet-$50$, we are able to match the FLOP reduction of state-of-the-art channel pruning results while achieving a $0.98\%$ higher accuracy. Further, we show that our pruned ResNet-$50$ network outperforms ResNet-$34$ and ResNet-$18$ networks, and that our pruned ResNet-$101$ outperforms ResNet-$50$.  
\end{abstract}

\section{Introduction}
In recent years, convolutional neural networks (CNNs) have become the predominant approach for a variety of computer vision tasks, e.g. image classification ~\cite{krizhevsky2012imagenet}, object detection \cite{girshick2014rich}, semantic segmentation \cite{long2015fully}, image captioning \cite{vinyals2015show}, and video analysis \cite{simonyan2014two}. 
Supported by the availability of high end modern GPUs and large scale labeled data sets \cite{deng2009imagenet}, the state-of-the-art CNN architectures have grown unprecedentedly large. For instance, a $152$-layer ResNet \cite{he2016deep} comprises more than $60$ million parameters and requires more than $20$ Giga floating-point-operations (FLOPs) when inferencing an image of $224 \times 224$ resolution. 

Such large networks admit large inference latency and require more space in memory. Therefore, Most of the standard CNN architectures have smaller versions. For example, 
ResNet has five standard sizes: ResNet-$152$, ResNet-$101$, ResNet-$50$, ResNet-$34$, and ResNet-$18$, where $X$ in ResNet-$X$ represent the number of convolutional layers. Similarly, MobileNet has three three standard sizes: MobileNet-$1.0$, MobileNet-$0.5$, MobileNet-$0.25$. All the three versions have the same number of convolutional layers, however uniformly across all the layers the number of filters in MobileNet-$0.5$ is half that of the MobileNet-$1.0$. 

In this work, we ask the following question: can we get an architecture more efficient than ResNet-$34$ by pruning channels of ResNet-$50$? In generality, can we get architectures more efficient than the expert-designed smaller versions of a large network by pruning channels of the large network in a data-driven way? By efficient architecture we mean an architecture which is better in terms of accuracy or latency/throughput/\#FLOPS/\#params. To the best of our knowledge none of the existing channel pruning works compare the pruned versions of a large network with the next available standard smaller network, for large data-sets like ImageNet. 

The recent works on model compression can be divided into four main categories, namely, quantization \cite{courbariaux2015binaryconnect, rastegari2016xnor, hubara2017quantized}, low rank factorization \cite{jaderberg2014speeding, lebedev2014speeding, tai2015convolutional}, sparse connections \cite{han2015deep, han2015learning}, and structured sparsity such as channel pruning \cite{wen2016learning, luo2017thinet, liu2017learning, zhuang2018discrimination}. Network quantization aims to reduce the model size and accelerate the inference by reducing precision of the network weights. Many of the modern computing devices support faster inference for the low precision networks. Low rank factorization approximates the convolutional kernels using tensor decomposition techniques. Sparse connections seeks to sparsify the network by pruning low importance weights in the model. 
We note that unstructured sparsity may lead to a reduction in the number of parameters, but typically will not accelerate the network due to its inherent need to access memory in a non-consecutive way.
In contrast, channel pruned network has exactly the same architecture and back-end implementation but with fewer filters and channels. Hence it immediately yields smaller memory footprint and faster inference than the original model without requiring any additional hardware or software support. Further, channel pruning is complementary to network quantization and it is known that both can be applied together to achieve higher compression than any method individually \cite{han2015deep}.

\begin{figure*}
 \begin{center}
	\includegraphics[width=1\linewidth]{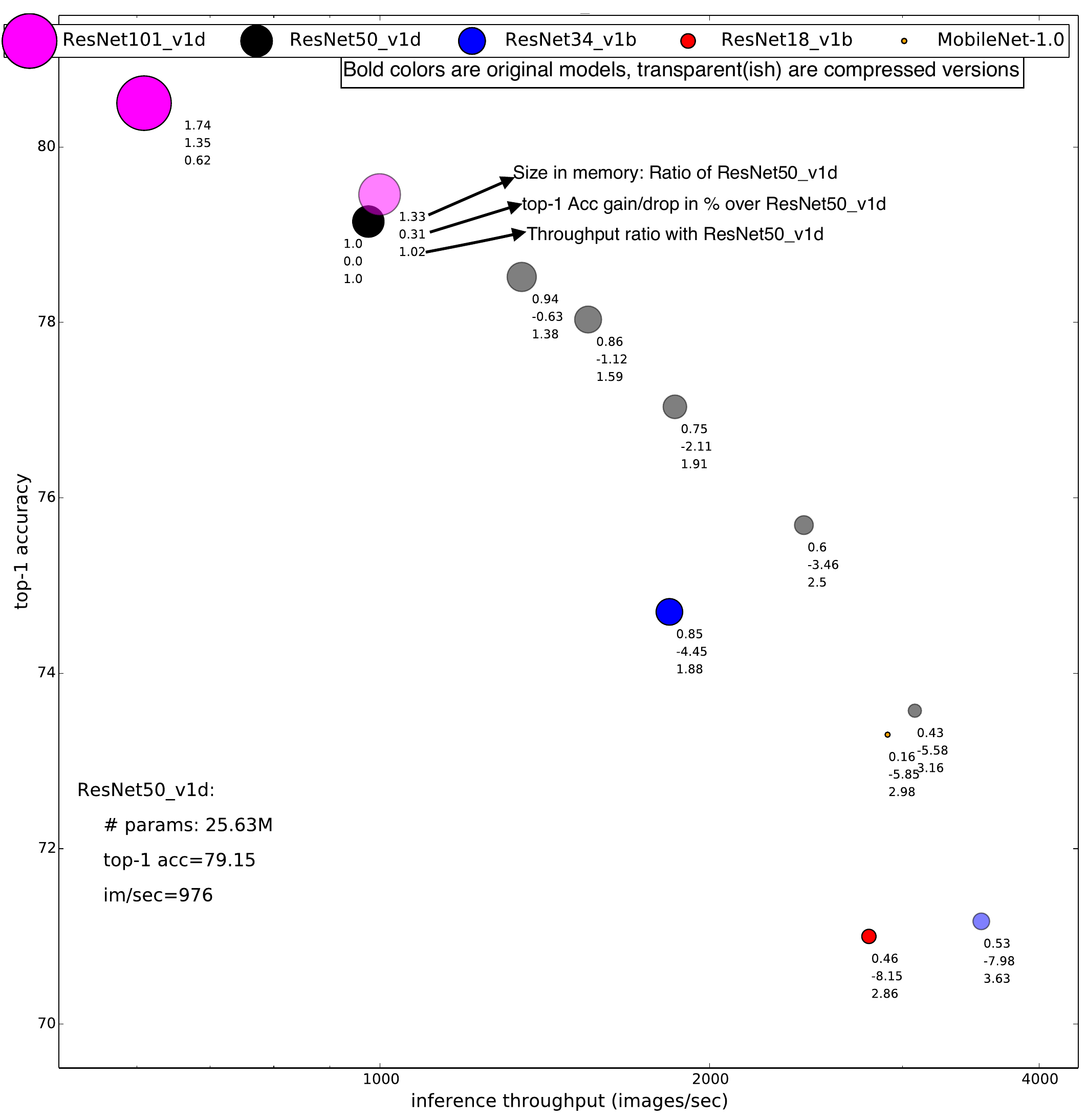}
\vspace{-1em}
 	\caption{The black circle corresponds to ResNet-50 model. It has 25.63 million parameters, 79.15\% top-1 accuracy on ImageNet dataset, and it gives a throughout of ~1060 images/sec for batch size of 64 on a NVIDIA V100 GPU. The other colored circles represent the other state-of-the-art models. The grey circles represent our PruneNets. The three numbers associated with each of these circles show the following: the first number gives size of the model (multiplicative w.r.t. ResNet-50), the second number gives loss in accuracy w.r.t. ResNet-50, and the third number gives multiplicative gain in throughput w.r.t. ResNet-50. 
 	}
	\label{fig:fig_main}
\end{center}
\end{figure*}

\section{Summary of Results}
\label{sec:summary}

In this work, we aim to investigate simple yet effective methods to perform channel pruning in the entire network across the layers, allowing the possibility of pruning more channels in one layer compared to others. This freedom gives our results a flavor of architecture search, as the resulting model's architecture is based on the dataset rather than an educated guess. We compare the pruned ResNet models with the next available standard smaller model. Figure~\ref{fig:fig_main} shows a scatter plot of various models. The x-axis represents throughput and the y-axis corresponds to accuracy. The grey colored dots represent our PruneNets, the networks obtained by pruning channels of ResNet-50. The base models and their training script have been taken from MXNET GluonCV GitHub repository. It can be seen that the PruneNets (pruned ResNet-$50$) significantly outperform ResNet-$34$, ResNet-$18$ and MobileNet-$1.0$. Pruned ResNet-$101$ (light pink color) outperforms ResNet-$50$ (black color). The light blue color dot on the extreme right represents pruned ResNet-$34$. 
The pruned ResNet models are available in GluonCV repository and their performance can be seen on accuracy-vs-throughput plot on its webpage \url{https://gluon-cv.mxnet.io/model_zoo/classification.html}.

In Figure~\ref{fig:fig_main}, we have pruned channels to maximize throughput on a batch-size of $64$ for NVIDIA V100 GPU. However, in general the objective of pruning channels can vary. We may want to minimize \#FLOPs or memory footprint or maximize throughput for a different batch-size on a different CPU/GPU machine. Our channel pruning approach takes the particular objective into consideration and accordingly gives different pruning pattern across the layers.  

\begin{figure}
 \begin{center}
	\includegraphics[width=0.9\linewidth]{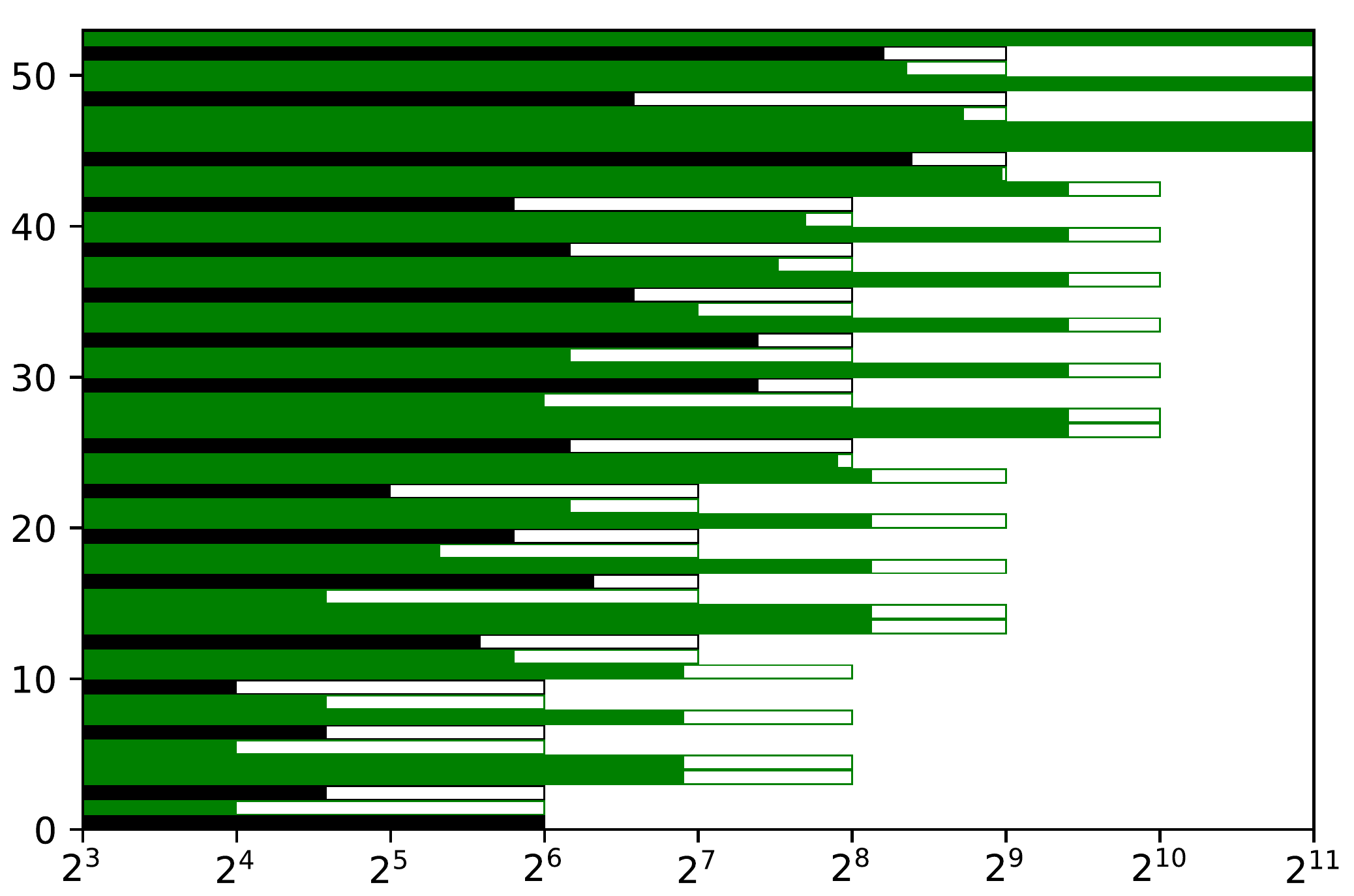}
	\put(-150,-10){{\small{size of output channels}}}
    \put(-210,20){ \rotatebox{90}{\small{index of convolution layer}}}
        \vspace{-1em}
	\caption{Pruning pattern achieved by our approach when the objective is to minimize \#FLOPs. Shallow layers are pruned aggresively. The black bars represent $3 \times 3$ convolution layers and the green bars represent $1 \times 1$ convolution layers. The white colored area in each bar represents pruning.     
	}
	\vspace{-1em}
	\label{fig:fig1}
\end{center}
\end{figure}

\begin{figure}
 \begin{center}
	\includegraphics[width=0.9\linewidth]{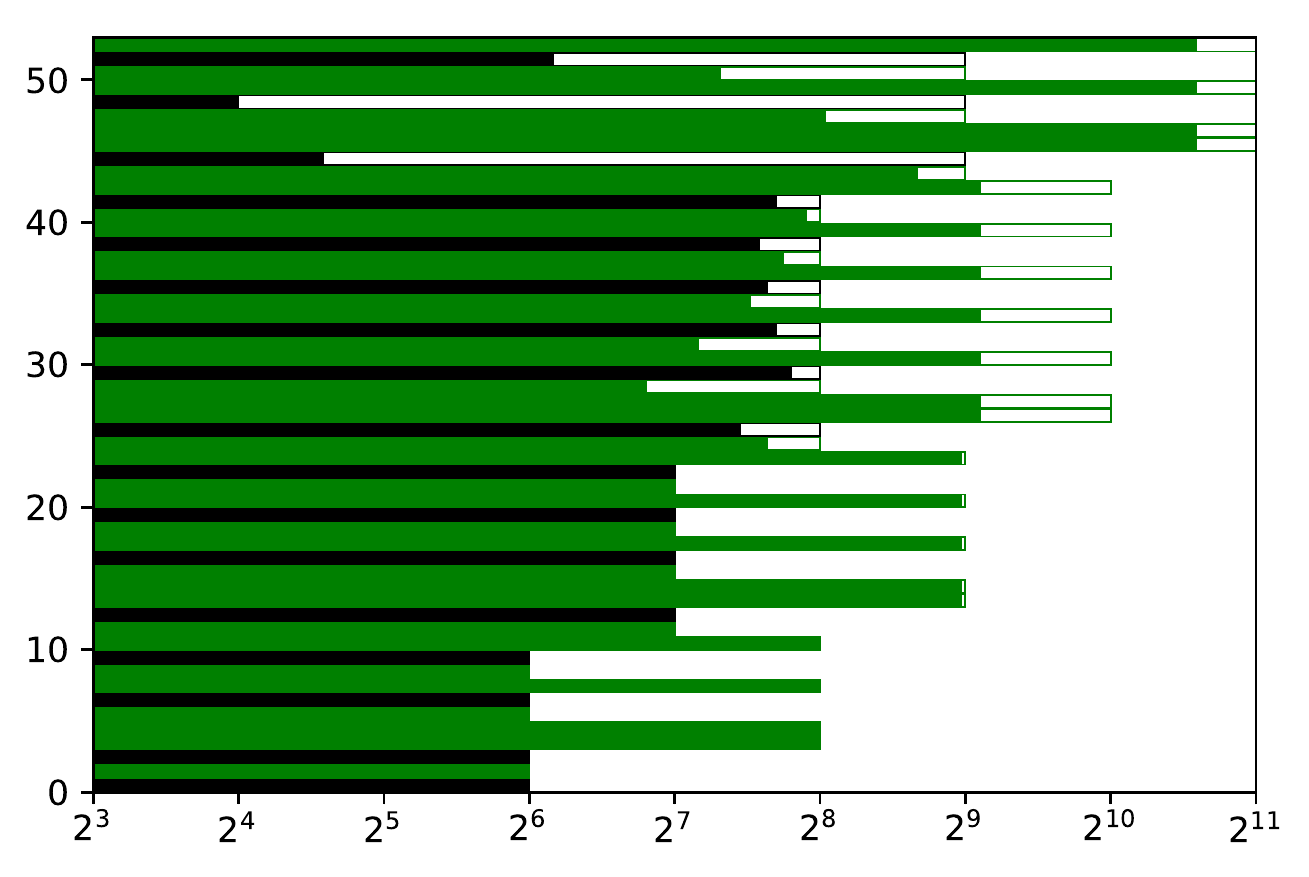}
	\put(-150,-10){{\small{size of output channels}}}
    \put(-210,20){ \rotatebox{90}{\small{index of convolution layer}}}
    \vspace{-1em}
	\caption{Pruning pattern achieved by our approach when the objective is to minimize \#Params. Deep layers are pruned aggresively. The black bars represent $3 \times 3$ convolution layers and the green bars represent $1 \times 1$ convolution layers. The white colored area in each bar represents pruning. 
	}
	\vspace{-1em}
	\label{fig:fig2}
\end{center}
\end{figure}

Figure \ref{fig:fig1} and Figure \ref{fig:fig2} show the channel pruning patterns achieved by our approach when the objective is to minimize \#FLOPs ($3.47 \times$ reduction) and when the objective is to minimize the number of parameters ($2.95 \times$ reduction) respectively. 
When the objective is to minimize the \#FLOPs, Figure \ref{fig:fig1}, the layers closer to the input are pruned aggressively whereas when the objective is to minimize the \#Params, Figure \ref{fig:fig2}, the layers closer to the output are pruned heavily. This very different pruning pattern depending upon the pruning objective can be understood by the fact that the layers closer to the input are more compute intensive (size of feature map is larger in the layers closer to the input) whereas the layers closer to the output are more parameter intensive (number of channels is larger in the layers closer to the output). The accuracy of these pruned networks are given in the eleventh row of Table \ref{tab:resnet50_1} and the seventh row of Table \ref{tab:resnet50_2} respectively in Section \ref{sec:experiment}. 

We note that most of the existing channel pruning works uniformly prune the channels across all the layers, or minimize the number of channels in the network ignoring the real objective (reducing FLOPs/Params) of channel pruning. 

In Section~\ref{sec:approach} we explain our channel pruning approach in detail. In Section~\ref{sec:experiment} we provide experiments on several ResNet architectures, on the ILSVRC-$12$ (ImageNet) dataset. Since the existing channel pruning works only report reduction in memory footprint and FLOPs count, for comparison purpose we restrict to these two objectives.  The resulting PruneNets achieve state-of-the-art performance when compared to other pruned models reported in the literature. In order to test the resulting model we assess not only its performance on a test set, but its applicability in other tasks. We use a common transfer learning task and measure how a pruned ResNet-$50$ model performs when fine tuned on the Caltech-$256$ dataset. The performance of the pruned model is comparable and in some settings even better than that of the original ResNet-$50$ model.

\section{Related works}
There have been a significant amount of work on compressing and accelerating deep CNNs. Most of these works fall primarily into one of the four categories: quantization, low rank factorization, sparse connections, and structured pruning. Besides compression methods, there have been a lot of work in architecture search for compute efficient networks \cite{liu2018darts, singh2019darc}. Recently, model compression techniques has also been applied on natural language processing models \cite{lan2019albert, khetan2020schubert}.  

Quantization approaches do not reduce the number of parameters but rather their precision, meaning the number of bits representing each parameters. This is done by quantizing the parameters to binary \cite{rastegari2016xnor, hubara2017quantized}, ternary \cite{zhu2016trained}, or 4 or 8 bits per parameter \cite{han2015deep, gong2014compressing}. Low rank factorization techniques exploit various low rank structures in the convolution parameters, such as a basis for the filters  \cite{jaderberg2014speeding}, or a tensor low rank decomposition \cite{lebedev2014speeding, tai2015convolutional}. We note that the practical compression and the acceleration achieved with these low-rank approaches is currently limited as the standard deep learning libraries do not support convolutional operation by low rank weight tensors.

Different methods have been explored to prune the model parameters. The early work by \cite{hanson1989comparing} performed magnitude based pruning and \cite{hassibi1993second} suggested Hessian matrix based approach to prune the network weights. More recently, \cite{han2015learning} introduced an iterative method to prune weights in deep architectures, together with an external compression by quantization and Huffman encoding \cite{han2015deep}. The network is pruned by removing low weight connections, followed by fine-tuning to recover its accuracy. Training with sparsity constraints has also been studied by  \cite{srinivas2017training, wen2016learning} to achieve higher compression. \cite{hu2016network} extends parameter pruning and seeks to identify the optimal number of parameters that should be pruned in each layer to minimize the loss in accuracy for a given budget on the pruning. They achieve this by using a measure of average percentage of zeros (APoZ) for each layer.

Existing works in channel pruning can be divided into two categories: training from scratch and pruning redundant channels in a pretrained network. The first type \cite{wen2016learning, alvarez2016learning, liu2017learning} trains the network with regularization terms aimed to induce sparsity along the channels. This approach has several shortcomings. First, by requiring full training it does not apply to situations where we already have a trained model and we aim to compress it in a lighter procedure involving either less computation or less data. Second, the sparsity inducing regularization can often insert bias that degrades performance. This issue can be somewhat mitigated by considering non-convex regularization or some iterative pruning approach (see e.g.\ \cite{carreira2018learning}); this fix however comes at the expense of a complicated system that requires a longer training time due to new hyper parameters to be optimized, or the need for multiple training rounds when pruning repeatedly without sparsity inducing regularization. Non-convex regularization typically has more than one parameter associated with it, informally determining how far away it is from L1 regularization.
Another issue comes up in networks that use Batch Norm \cite{liu2017learning}. In all recent architectures a batch-norm layer is present after every convolution. The normalization operation done by this layer can completely modify the effect of a regularization term, as the magnitude of the parameters does not change the output of the model. For example, \cite{HofferBGS18} show that the $\ell_2$ regularization term's effect can be mimicked with a learning rate schedule, meaning that the `weight-decay' parameter helps more as a learning rate scheduler as opposed to a regularization parameter. This result along with additional papers cited within suggest that a naive combination of sparsity inducing regularization combined with batch-norm may not have the desired effect. 

The second type of channel pruning works that start with a pretrained model can be further sub-categorized into two. First sub-category of works prune channels uniformly in each layer using reconstruction based approaches \cite{luo2017thinet, he2017channel}. The high level mode of operation is to obtain a sample of inputs and outputs of a convolutional layer, then learn a layer with lesser input channels that produce approximately the same output.  
The measure of approximation is typically in $\ell_2$ norm but can be more sophisticated, see e.g. \cite{zhuang2018discrimination}. The main drawback of these methods comes from the fact they handle convolutional layers one at a time. As such, they cannot detect correlations occurring across layers, and more important, it is hard to adapt these methods to prune one layer more aggressively than another. It follows that these works are inherently limited as they prune the same fraction of channels across all the layers. Our methods can in fact be seen as complimentary to these results, as we provide a light-weight scheme that among others, discovers the extent by which the layers should be pruned.
The second sub-category of works prune channels in a pretrained model based on the network parameters and their basic statistics. \cite{li2016pruning} prunes channels based on the $\ell_1$ norm of the filters, and \cite{hu2016network} defines redundancy of neurons based on their activations and prunes the more redundant ones. In contrast to these methods, our method is optimization driven. Instead of identifying redundancy of channels based on parameter values of the pretrained model we optimize over parameters which correlate with the redundancy of the channels. Therefore our method encompasses these existing methods and improves upon them. 


Our paper presents a channel pruning technique that enjoys the global view and precise pruning of the full training techniques, while requiring a short training time for the pruning procedure, similarly to the techniques that prune based on basic statistics of the weights. In contrast to the papers described above, the importance of channels is based on the motivating example of sparse linear regression, takes into account the affects of BatchNorm, and most importantly, is based on a short (handful of epochs) training procedure rather than the model weights or a few basic statistics of the data. Our claim that this technique achieves a more accurate importance ranking for the channels is backed up by our experiments achieving state-of-the-art results on ResNet architectures on the ImageNet dataset.

\section{Approach}
\label{sec:approach}
\newcommand{\B}{\mathcal{B}}
We begin by formally characterizing a CNN and its parameters.  For readability we assume the structure of Convolution-BatchNorm-Activation and a ReLU activation. We note that our framework applies without this assumption, meaning for any order and for any activation type. Let $L$ denote the number of convolution operators in the network, and for $\ell \in [L] = \{1,\ldots,L\}$ let $W_\ell \in \R^{n_\ell \times m_\ell \times k_\ell \times k_\ell}$ denote the filter weights of the $\ell$-th convolution. Here, $n_\ell$ and $m_\ell$ represent the number of output and input channels respectively, and  $k_\ell \times k_\ell$ is the size of each filter. A BatchNorm (BN) layer has the same input and output size. Denote by  $z^{(in)} \in \mathbb{R}^{n \times r \times r}$ and $z^{(out)} \in \mathbb{R}^{n \times r \times r}$ its input and output respectively. Here, the $z$ variables contain $n$ channels, each assumed to have $r \times r$ features. This assumption is purely for readability. Our framework works for 1d or 3d convolution as well, with any feature map size.
Let $\B$ denote the current mini-batch, a standard BN layer performs the following affine transformation for each $i$-th feature map $z_{i} \in \mathbb{R}^{r \times r}$, for $i \in \{1,2,\cdots, n\}$.
\begin{equation} 
    \hat{z}_{i} = \frac{z^{(in)}_{i} - \mu_{\B_i}}{\sqrt{\sigma^2_{\B_i} + \epsilon}}; \;\;\;\; z^{(out)}_{i} = \gamma_{i} \hat{z}_{i} + \beta_{i}\,,
    \label{eq:def_BN}
\end{equation}
where $\mu_{\B_i}$ is the scalar mean of the entire $i$-th feature map over the mini-batch $\mathcal{B}$, and $\sigma^2_{\B_i}$ is the scalar variance. Although individual numbers within the feature map $\hat{z}_i$ do not have a zero mean and unit variance, each feature map $\hat{z}_i$, when considered as a $r^2$-dimensional vector, has unit norm and sums to zero, in expectation. With this in mind, we say that $\gamma_i^2$ controls the variance of the $i$-th output channel $z^{(out)}_{i}$, and $\beta_i$ is its mean. 

When doing inference, the BN layer behaves slightly differently. Rather than taking the mini-batch statistics $(\mu_{\B_i}, \sigma^2_{\B_i})$, it uses their global counterparts $(\mu, \sigma^2)$ obtained not from a single mini-batch but from the entire dataset (or sufficiently large sample).
Since each convolution is followed by a BN layer, we let $\gamma_{\ell,i}$, $\beta_{\ell, i}$ represent the $i$-th scale and bias parameters of the $\ell$-th convolution layer. 

For the entire network, let $W \equiv \{W_{\ell}\}_{\{1,2,\cdots, L\}}$ denote the set of all convolution parameters, and let $\gamma=\{\gamma_{\ell,i}, \beta_{\ell,i}\}_{\ell,i}$ denote the parameters of the BN layers following the convolutions. Denote all other network parameters by $F$. This could include fully connected layers etc. 

\subsection{Global Importance Score}
We are now ready to explain our approach and the motivation leading to it. We first observe that if we ignore the effect of the activations, the $i$-th input channel of the $\ell$-th convolution layer has a variance of $\gamma_{\ell-1,i}^2$. 
Therefore, the contribution of the $i$-th input to the variance of the $j$-th output in the $\ell$-th convolution layer is
\begin{equation} \label{eq:important1}
\gamma_{\ell-1,i}^2 \|W_{\ell,i,j}\|_2^2\,,
\end{equation}
where $W_{\ell,i,j}$ is the filter in the $\ell$-th convolution layer corresponding to the $i$-th input and the $j$-th output. Indeed if the filters and input feature maps were of dimension $1$, then the convolution would be reduced to a linear function. 
In this simplified case where the convolution is reduced to a linear function, if we wanted to ignore one of the inputs and minimize the squared distance of the output change for the $j$-th output channel, the importance of the inputs would exactly be determined by the importance score in Equation~\eqref{eq:important1}. 
When considering all outputs simultaneously the score becomes 
\begin{equation} \label{eq:important}
\gamma_{\ell-1,i}^2 \sum_{j=1}^{n_\ell} \|W_{\ell,i,j}\|_2^2\,.
\end{equation}
We let this score be the global importance score of the $i$-th input channel to the $\ell$-th convolution layer.
Now, since the ReLU activation is linear, and as detailed before, when applying sparsity regularization we do not modify the convolution weights $W$, rather we multiply the convolution weights with a scalar such that $\sum_{j=1}^{n_\ell} \|W_{\ell,i,j}\|_2^2=1$ and shift that scalar into $\gamma_{\ell-1,i}$ and $\beta_{\ell-1, i}$. For this reason, when we discuss the details of our implementation we will assume that the importance score is simply $\gamma_{\ell-1,i}$.

\subsection{Regularization and Importance Scores}

Given this Global Importance Score, a naive approach for pruning channels of a pretrained network would be to rank all the channels according to their score and prune the least important ones. However, a sophisticated approach would be to train a model while regularizing the cost associated with each channel and then prune the low importance channels.

Let's proceed to define our optimization objective. First, denote by 
\begin{eqnarray}\label{eq:eq2}
\mathcal{L}\Big(f\big(x; \{W, F, \gamma, \beta\}; \{\mu_{\B}, \sigma^2_{\B}\}\big), \;\;y\Big)\,,
\end{eqnarray}
the loss of the network during a standard training. The most straightforward objective when considering channel pruning is
\begin{eqnarray*} 
\eqref{eq:eq2} + \lambda  \sum_{\ell>1,i}\alpha_{\ell} \mathbf{1}[\gamma_{\ell-1,  i}> 0]
\end{eqnarray*}
Here, a channel with $\gamma_{\ell,i}=0$ is considered pruned even if $\beta_{\ell,i} \neq 0$, since the bias term can be simulated by modifying the other network parameters. We let $\alpha_\ell$ denote the cost associated with each input channel of the $\ell$-th convolutional layer. 
The value of the $\alpha$ costs are determined by our objective. For example, for FLOPs minimization we compute the number of FLOPs needed for the full network, and for the network after pruning an output of the $(\ell-1)$-th layer which is also an input to the $\ell$-th layer. This difference determines the scale of $\alpha_\ell$. The objective of channel pruning can be any of the following: minimize the model size (\# params), reduce the \# FLOPs required for inferencing an image, or minimize latency on a given hardware (GPU/CPU). We choose the costs $\alpha_\ell$ according to the objective.

Since $\ell_0$ norm is non-smooth and has zero gradient almost everywhere, we consider the standard continuous relaxation of it and use the $\ell_1$ norm which is well known to induce sparsity. Therefore, we seek to minimize the following objective function. We note that there are papers that use various non-convex regularization. We postpone exploring the benefit of those to future works, so as to minimize the number of unknowns.
\begin{eqnarray} 
\label{eq:main_obj}
\mathcal{L}\Big(f\big(x; \{W, F, \gamma, \beta\}; \{\mu_{\B}, \sigma^2_{\B}\}\big), \;\;y\Big) \nonumber\\
\;+\;  \lambda  \sum_{\ell,i}\alpha_{\ell} |\gamma_{\ell,i}|\,.
\end{eqnarray}

\subsection{Optimize with BatchNorm off}
An immediate approach would be to optimize the regularized objective in \eqref{eq:main_obj} by training all the model parameters $W, F, \gamma, \beta$ with random initialization. Recall $(\mu_{\B}, \sigma^2_{\B})$ are the mini-batch mean and variance values, not the trainable model parameters. 
However, a closer investigation of BatchNorm layer, \eqref{eq:def_BN}, reveals that in doing so the impact of regularizing $\gamma_{\ell, i}$ would be nullified by the normalization operation in the BatchNorm layer. Particularly, the regularization term $\sum_{\ell,i}\alpha_{\ell} |\gamma_{\ell,i}|$ can be pushed to any arbitrary  positive value $\epsilon > 0$ without changing the network output and the loss function $\mathcal{L}(\cdot)$. It can be seen that for any given scalar $\tau > 0$ there exists $\widetilde\gamma = \tau \gamma$, $\widetilde\beta = \tau\beta$, $\widetilde\mu_{\B} = \tau\mu_{\B}$, and $\widetilde\sigma^2_{\B} = \tau^2 \sigma^2_{\B}$, such that
\begin{eqnarray*}
&&\mathcal{L}\Big(f\big(x; \{W, F, \gamma, \beta\}; \{\mu_{\B}, \sigma^2_{\B}\}\big), \;\;y\Big) \nonumber\\
& = & \mathcal{L}\Big(f\big(x; \{W, F, \widetilde\gamma, \widetilde\beta\}; \{\widetilde\mu_{\B}, \widetilde\sigma^2_{\B}\}\big), \;\;y\Big)\,.
\end{eqnarray*}
Note that when the $\gamma$ and $\beta$ are reduced by a factor of $\tau$ then by definition, \eqref{eq:def_BN}, the mean and variance of feature maps change to $\widetilde\mu_{\B}$ and $\widetilde\sigma^2_{\B}$.
Therefore, we propose optimizing the regularized objective \eqref{eq:main_obj} by turning the BatchNorm layer off. We do so by fixing the BatchNorm mean and variance values $(\mu_{\B}, \sigma^2_{\B})$ to their global counterparts $(\mu, \sigma^2)$. Note that during inference also $(\mu_{\B}, \sigma^2_{\B})$ are set to $(\mu, \sigma^2)$. Further, the convolutional weight parameters $W$ also have the same impact on the regularization term; $\gamma$ and $\beta$ can be reduced by a factor of $\tau$ while keeping the network output same by increasing $W$ by a factor of $1/\tau$. Therefore, we start with a pretrained model and fix the convolution weight parameters $W$ to their value $W^*$ in the pre-trained network. We minimize the following objective to induce sparsity in $\gamma$. 
\begin{eqnarray} \label{eq:phase1.1}
\mathcal{L}\Big(f_{(W^*,\mu,\sigma^2)}\big(x; \{F, \gamma, \beta\}; \big), \;\;y\Big) \nonumber\\
\;+\; \lambda  \sum_{\ell,i}\alpha_{\ell} |\gamma_{\ell,i}|\,.
\end{eqnarray}
Note that the objective is only a function of BatchNorm scale parameters $\gamma, \beta$ and the fully connected layer parameters $F$. Since $F$ is typically a single fully connected layer, the overall number of parameters to be optimized is small and a single epoch suffices.

Note that \cite{liu2017learning} follows the approach of optimizing the objective in \eqref{eq:main_obj} by training all the model parameters $W, F, \gamma, \beta$ with random initialization. However, as explained above this approach nullifies the effect of regularizing $\gamma$, and requires a long time to converge due to the large number of parameters in $W$.



\subsection{Step-wise Pruning of Channels}
The step of optimizing \eqref{eq:phase1.1} leaves us with a ranking of the channels based on their Importance Score $\gamma$. We prune sufficiently many channels to reduce the total cost $C= \sum_{\ell, i} \alpha_\ell$ to $\eta C$, for a given fraction $\eta$. By prune we mean we set their $\gamma$ value to zero if it wasn't already zero, and move on to the next phase. 

In the next phase we fix the bias suffered by the sparsity inducing regularization and optimize
\begin{eqnarray} \label{eq:phase1.2}
\mathcal{L}\Big(f_{W^*}\big(x; \{F, \gamma, \beta\}; \{\mu_\B, \sigma^2_\B\}\big), \;\;y\Big) ,
\end{eqnarray}
the same objective as in~\eqref{eq:phase1.1} but without the regularization term. In this step, we use the mini-batch mean and variance values $(\mu_B, \sigma^2_B)$ for BatchNorm layers. Also, we compute the updated values of their global counterparts $(\mu, \sigma)$ which are used in the next iteration of the regularization, \eqref{eq:phase1.1}.
Here, we abused notation and use $\gamma$ to denote the set of non-pruned weights. The pruned weights are fixed as zeros. As before, since the number of parameters here is small, a single epoch typically suffices to solve the optimization problem.

Finally, we alternately repeat optimizing objectives~\eqref{eq:phase1.1} and \eqref{eq:phase1.2} for $T$ times. In each step we reduce the cost by $(\eta/T)C$ for a desired fraction of cost reduction $\eta$. We optimize the hyper-parameters associated with these two steps in a way that minimizes the loss of Equation~\eqref{eq:phase1.2}. Since the overall procedure is lightweight, exploring hyper-parameters is much cheaper compared to those relevant to a full training job. 

Once we obtained the final pruned model, we fine tune all of its weights, including $W$, by optimizing objective~\eqref{eq:eq2}, with the appropriate $\gamma$ values being fixed as zeros.
The following Algorithm \ref{algo:algo1} summarizes our approach. 

\renewcommand{\algorithmicrequire}{\textbf{Input:}}
\renewcommand{\algorithmicensure}{\textbf{Output:}}

\begin{center}
\begin{algorithm} 
\caption{Global Channel Pruning (GCP)}
\begin{algorithmic} \label{algo:algo1}
    \REQUIRE A CNN architecture, Minimization objective (\# FLOPs/ \# Params), target fraction $\eta$, $T$: number of iterations.
	\ENSURE A pruned CNN model.
    	\STATE \hspace{-1em} {Train the network using loss \eqref{eq:eq2}.}
	\STATE Let $W^*$ be the filter weights of the trained network. 
	\STATE \hspace{-1em}\textbf{Repeat $T$ times:}
	\STATE Fix $W^*$ and the BN mean and variance parameters to their global counterparts.
	\STATE Train the network with regularization on $\gamma$'s using loss \eqref{eq:phase1.1}. 
	\STATE Rank the channels according to their global importance score $\gamma$.
	\STATE Prune channels according to their importance score to achieve $\eta/T$ reduction in the target objective value (\# FLOPs/ \# Params)
	\STATE Train the network without regularization term on $\gamma$ using loss \eqref{eq:phase1.2}
\STATE \hspace{-1em}\textbf{Retrain the pruned network including filter weights $W^*$ using loss \eqref{eq:eq2}.}
\end{algorithmic}
\end{algorithm}
\end{center}


\section{Experiments} \label{sec:experiment}
In this section, we empirically evaluate the performance of our Global Channel Pruning (GCP) algorithm. The GCP operates on a pretrained model and requires two inputs: the minimization objective, for e.g. minimize \# FLOPs, and the target fraction, $\eta$. The target fraction $\eta$ is the ratio of the desired objective value in the pruned network and the objective value in the original network. Note that we chose the regularization scale $\alpha_\ell$ for each channel $\ell \in \{1,2,\cdots,L\}$ according to the objective we seek to minimize. We  refer GCP by GCP-p when $\alpha_\ell$ are chosen to minimize the \# params and by GCP-f when $\alpha_\ell$ are chosen to minimize \# FLOPs. We refer GCP by GCP-$\ell$ when $\alpha_\ell$ are chosen to minimize the latency. Our method can also be used to maximize throughput as it is the inverse of latency.  

We evaluate GCP on state-of-the-art ResNet models for ILSVRC-$12$ (ImageNet) and Caltech-$256$ datasets. We compare performance of GCP against several well known channel pruning methods, including Discrimination Aware Channel Pruning (DCP) \cite{zhuang2018discrimination}, ThiNet\cite{luo2017thinet}, and Channel Pruning (CP) \cite{he2017channel}. To the best of our knowledge, these are the state-of-the-art works that report channel pruning results on ResNet models for ImageNet dataset. These works set a target fraction $\eta$ for pruning channels and uniformly prune the $\eta$ fraction of channels from each layer. However, since the true objective of channel pruning is to reduce the  \# params and \# FLOPs, they report these numbers as well. 
To fairly compare with these works we run GCP-f with the target fraction $\eta$ equal to the FLOPs reduction reported in these works, and run GCP-p with $\eta$ set to the parameter reduction reported in these works.

Besides, to investigate the effect of the non-uniform pruning achieved by GCP, we study the uniform pruning approach wherein we prune each layer uniformly. We prune $\eta$ fraction of channels from each layer in the pretrained model and retrain the pruned model. Though the DCP/ThiNet/CP prune uniformly across the layers, they are significantly distinct than the naive uniform pruning approach. They employ variants of reconstruction based approaches to identify which channels to prune in the each layer as against randomly pruning $\eta$ fraction of channels. 

\subsection{Implementation details}
We implement GCP on the MXNET deep learning framework. For training the original ResNet networks, we use the training script provided in the MXNET library- GluonCV. 

ResNet is a residual network architecture. It has many skip connections which require that the size of the output channels be same for the layers connected by the skip connections. If the size of the output channels is not same, then it requires sparse additions in the residual addition layer instead of the standard element-wise addition. Since the standard deep learning libraries, including MXNET, are highly optimized to perform the standard convolution operations faster, such sparse addition layers tend to slow down the network. Therefore, we restrict the pruning of layers connected by the skip connection to be identical. This is done by multiplying (element-wise) a mask parameter $\tilde{\gamma}_a$ to all the $\gamma_{\ell}$ parameters of the layers which need to be pruned identically (due to the dependency enforced by the skip connection). Instead of regularizing $\gamma_\ell$'s for these layers we regularize $\tilde{\gamma}_a$, which ensures that these layers are pruned identically. Further, we use proximal gradient descent \cite{parikh2014proximal} to optimize the $\gamma$ parameters during the regularization step when we need to sparsify $\gamma$. 

\subsection{Comparisons on ILSVRC-$\mathbf{12}$}
ILSVRC-$12$ (ImageNet) contains $1.28$ million training images and $50$ thousand testing images for $1000$ classes. This is one of the most widely used datasets for evaluating performance of CNNs. Further, networks trained on ImageNet are commonly used for transfer learning on smaller datasets such as Caltech-256. We show effectiveness of GCP on three different ResNet architectures, namely ResNet-$18$, ResNet-$50$ and ResNet-$101$. On all the three networks, in the high pruning regime, GCP outperforms the baselines models with a significant margin. In the low pruning regime, its performance is comparable to the baseline models. In all the settings, it performs better than the naive uniform pruning. 

\subsubsection{Resnet-$\mathbf{18}$}
Table \ref{tab:resnet18_1} gives the results of channel pruning of ResNet-$18$ for various FLOPs reduction targets. For example, the numbers in the sixth row correspond to the results of GCP-f for \# FLOPs minimization with target fraction $\eta=1/1.87$. It returns a pruned model that has $1/1.19$ fewer parameters, $1.70\%$ higher top-$1$ and $1.06\%$ higher top-$5$ error rate than the original pretrained model. Compared to this, DCP achieves the same \# FLOPs reduction at $2.29\%$ increase in top-$1$ error rate. Note that since GCP-f prunes channels non-uniformly, its parameter reduction is not same as its FLOPs reduction unlike the case of naive Uniform pruning and the DCP approach which prunes channels uniformly.

Table \ref{tab:resnet18_2} gives similar results for the case of GCP-p applied on ResNet-$18$ for minimizing the model parameters. Again, in the high pruning regime, GCP-p significantly outperforms the DCP \cite{zhuang2018discrimination}.  
\setlength{\tabcolsep}{3pt}
\begin{table}
	\begin{center}
		\begin{tabular}{l||cccc}
			\hline
		      ResNet-18 & \# Params. $\downarrow$ &  \# FLOPs $\downarrow$ & Top-1  & Top-5 \\
		      \hline\hline
		      DCP & $1.39 \times$ & $1.37 \times$ & $\mathbf{+0.43}$ & $\mathbf{+0.12}$ \\
		      Uniform & $1.42 \times$ & $1.35 \times$& $+1.31$& $+0.67$ \\
		      GCP-f & $1.03 \times$ & $1.36 \times$ & $+0.45$ & $+0.32$ \\
			\hline \hline      
		      DCP & $1.89 \times$ & $1.85 \times$ & $+2.29$ & $+1.38$ \\
  		      Uniform & $1.92 \times$ & $1.85 \times$& $+3.01$& $+1.90$ \\
		      GCP-f & $1.19 \times$ & $1.87 \times$& $\mathbf{+1.70}$  & $\mathbf{+1.06}$ \\
			\hline \hline      
		      DCP & $2.92 \times$ & $2.79 \times$ & $+5.52$ & $+3.30$ \\
		      Uniform & $2.76 \times$ & $2.79 \times$& $+6.75$& $+4.2$ \\
		      GCP-f & $1.53 \times$& $2.81 \times$ &$\mathbf{+4.34}$ & $\mathbf{+2.70}$ \\
		\end{tabular}
	\end{center}
\vspace{-1.5em}
\caption{Comparisons on ILSVRC-$12$ for ResNet-$18$. The top-$1$ and top-$5$ error  \% of the pre-trained model are $29.21$ and $10.13$ respectively. The proposed method GCP-f incurs $1.18\%$ lower error rate than the DCP method for the same $2.81 \times$ reduction in \# FLOPs.}
\label{tab:resnet18_1}
\end{table}

\setlength{\tabcolsep}{3pt}
\begin{table}
	\begin{center}
		\begin{tabular}{l||cccc}
			\hline
			  ResNet-18 & \# Params. $\downarrow$ &  \# FLOPs $\downarrow$ & Top-1  & Top-5 \\
		      \hline\hline
		      DCP & $1.39 \times$ & $1.37 \times$ & $\mathbf{+0.43}$ & $\mathbf{+0.12}$ \\
		      GCP-p & $1.41 \times$ & $1.09 \times$ & $+0.71$ & $+0.31$ \\
			\hline \hline      
		    DCP & $1.89 \times$ & $1.85 \times$ & $+2.29$ & $+1.38$ \\
		      GCP-p & $1.89 \times$ & $1.21 \times$& $\mathbf{+1.84}$  & $\mathbf{+1.18}$ \\
			\hline \hline      
		      DCP & $2.92 \times$ & $2.79 \times$ & $+5.52$ & $+3.30$ \\
		      GCP-p & $2.97 \times$& $1.38 \times$ &$\mathbf{+5.21}$ & $\mathbf{+2.84}$ \\
		\end{tabular}
	\end{center}
	\vspace{-1.5em}
\caption{Comparisons on ILSVRC-$12$ for ResNet-$18$. The top-$1$ and top-$5$ error  \% of the pre-trained model are $29.21$ and $10.13$ respectively. In the high pruning regime, the proposed method GCP-p outperforms the DCP.}
\label{tab:resnet18_2}
\end{table}

\subsubsection{ResNet-$\mathbf{50}$}
Table \ref{tab:resnet50_1} gives the results of channel pruning of ResNet-$50$ for various FLOPs reduction targets. For example, the numbers in the eleventh row correspond to the results of GCP-f for \# FLOPs minimization with target fraction $\eta=1/3.47$. It returns a pruned model that has $1/2.18$ fewer parameters, $2.28\%$ higher top-$1$ and $1.35\%$ higher top-$5$ error rate than the original pretrained model. Compared to this, DCP achieves the same \# FLOPs reduction at $3.26\%$ increase in top-$1$ error rate. Further, the last four rows show the gain of GCP-f over uniform pruning in the extreme setting of $5 \times$ and $10 \times$ FLOPs reduction. 

Table \ref{tab:resnet50_2} gives similar results for the case of GCP-p applied on ResNet-$50$ for minimizing the model parameters. Again, in the high pruning regime, GCP-p significantly outperforms the DCP \cite{zhuang2018discrimination}.  

Figure ~\ref{fig:fig_main} in Section \ref{sec:summary} shows the results of GCP-$\ell$ applied on ResNet-$50$ for maximizing the throughput for a batch-size of $64$ on a NVIDIA V100 GPU. As discussed in Section \ref{sec:summary}, PruneNets significantly outperform the smaller networks- ResNets-$34$, ResNet-$18$ and MobileNet $1.0$. 

\setlength{\tabcolsep}{3pt}
\begin{table}
	\begin{center}
		\begin{tabular}{l||cccc}
			\hline
		      ResNet-50 & \# Params. $\downarrow$ &  \# FLOPs $\downarrow$ & Top-1  & Top-5 \\
		      \hline\hline
		      DCP & $1.51 \times$ & $1.56 \times$ & $\mathbf{-0.39}$ & $\mathbf{-0.14}$ \\
		      Uniform & $1.57 \times$ & $1.55 \times$& $+0.57$& $+0.18$ \\
  		      GCP-f & $1.12 \times$ & $1.55 \times$ & $+0.38$& $+0.17$ \\
			\hline \hline      
			ThiNet & $2.06 \times $& $2.25 \times$ & $+1.87$ & $+1.12$ \\
		    DCP & $2.06 \times$ & $2.25 \times$ & $+1.06$ & $+0.61$ \\
		    CP &  \_ & $2 \times$ & \_ & $+1.40$ \\
		    Uniform & $2.26 \times$ & $2.24 \times$& $+1.37$& $+0.74$ \\
		    GCP-f & $1.32 \times$ &$2.24 \times$ &$\mathbf{+1.02}$ &$\mathbf{+0.56}$ \\
			\hline \hline      
		    DCP & $2.94 \times$ & $3.47 \times$ & $+3.26$ & $+1.80$ \\
	        Uniform & $3.51 \times$ & $3.46 \times$& $+3.10$& $+2.14$ \\
		  GCP-f &$2.18 \times$ & $3.47 \times$& $\mathbf{+2.28}$ &$\mathbf{+1.35}$  \\
		      \hline\hline
	        Uniform & $5 \times$ & $5 \times$& $+5.08$& $+2.8$ \\
		      GCP-f &$3.09 \times$ & $5 \times$& $\mathbf{+3.86}$ &$\mathbf{+2.06}$  \\
		      \hline\hline
	        Uniform & $10 \times$ & $10 \times$& $+10.05$& $+5.89$ \\
		      GCP-f &$5.75 \times$ & $10 \times$& $\mathbf{+8.16}$ &$\mathbf{+4.89}$  \\		      
		\end{tabular}
	\end{center}
	\vspace{-1.5em}
\caption{Comparisons on ILSVRC-$12$ for ResNet-$50$. The top-$1$ and top-$5$ error  \% of the pre-trained model are $22.81$ and $6.47$ respectively. ``\_" denotes that the results are not reported.}
\label{tab:resnet50_1}
\end{table}

\setlength{\tabcolsep}{3pt}
\begin{table}
	\begin{center}
		\begin{tabular}{l||cccc}
			\hline
		      ResNet-50 & \# Params. $\downarrow$ &  \# FLOPs $\downarrow$ & Top-1  & Top-5 \\
		      \hline\hline
		      DCP & $1.51 \times$ & $1.56 \times$ & $\mathbf{-0.39}$ & $\mathbf{-0.14}$ \\
		      GCP-p & $1.51 \times$ & $1.13 \times$ & $-0.05$& $-0.19$ \\
			\hline \hline      
			ThiNet & $2.06 \times $& $2.25 \times$ & $+1.87$ & $+1.12$ \\
		    DCP & $2.06 \times$ & $2.25 \times$ & $+1.06$ & $+0.61$ \\
		    GCP-p & $2.08 \times$ & $1.32 \times$ & $\mathbf{+0.55}$ &$\mathbf{+0.20}$ \\
			\hline \hline      
		    DCP & $2.94 \times$ & $3.47 \times$ & $+3.26$ & $+1.80$ \\
		    GCP-p &$2.95 \times$ & $1.53 \times$& $\mathbf{+1.52}$ &$\mathbf{+0.65}$  \\
		\end{tabular}
	\end{center}
	\vspace{-1.5em}
\caption{Comparisons on ILSVRC-$12$ for ResNet-$50$. The top-$1$ and top-$5$ error  \% of the pre-trained model are $22.81$ and $6.47$ respectively. }
\label{tab:resnet50_2}
\end{table}

\subsubsection{ResNet-$\mathbf{101}$}
Table \ref{tab:resnet101_1} gives the results of channel pruning of ResNet-$101$ network for various FLOPs reduction targets. 
For both the FLOPs reduction target values of $\eta = 1/2$ and $\eta = 1/3.35$, GCP-f outperforms the naive uniform pruning approach. 
Figure ~\ref{fig:fig_main} in Section \ref{sec:summary} shows the results of GCP-$\ell$ applied on ResNet-$101$ for maximizing the throughput for a batch-size of $64$ on a NVIDIA V100 GPU, and results in a network both faster and more accurate than ResNet-$50$.
\setlength{\tabcolsep}{3pt}
\begin{table}
	\begin{center}
		\begin{tabular}{l||cccc}
			\hline
		      ResNet-101 & \# Params. $\downarrow$ &  \# FLOPs $\downarrow$ & Top-1  & Top-5 \\
		      \hline\hline
		      Uniform & $2.03 \times$ & $2 \times$& $+0.95$& $+0.34$ \\
		      GCP-f & $1.56 \times$ & $2 \times$ & $\mathbf{+0.68}$ & $\mathbf{+0.18}$ \\
			\hline \hline      
  		      Uniform & $3.59 \times$ & $3.35 \times$ & $+2.35$ & $+0.75$ \\
		      GCP-f & $2.12 \times$ & $3.35 \times$& $\mathbf{+2.11}$  & $\mathbf{+0.54}$ \\
		\end{tabular}
	\end{center}
	\vspace{-1.5em}
\caption{Comparisons on ILSVRC-$12$ for ResNet-$101$. The top-$1$ and top-$5$ error \% of the pre-trained model are $19.57$ and $4.94$ respectively.}
\label{tab:resnet101_1}
\end{table}

\subsection{Comparisons on Caltech-$\mathbf{256}$}
In the following we show that the pruned ResNet models when used for transfer learning on smaller dataset such as Caltech-$256$ performs comparable to the original model. Caltech-$256$ contains $30$ thousand images for $256$ classes. We resize the top fully connected layer of the pruned ResNet network to match the output size of $256$ classes and randomly initialize its weights. For training and testing we follow the standard protocol. We sample 60 images from each class as the training set, and the rest for the test set.  We fine-tune the network for $60$ epoch with a learning rate of $0.01$ with cosine learning scheduler. Table~\ref{tab:caltech256} compares the results obtained from various pruned versions of ResNet-$50$ with those obtained via the unpruned ResNet-$50$. For the lightest pruned version of ResNet-50 ($1.55 \times$ FLOPs reduction), the performance actually improved. It remains comparable even with $2.24 \times$ FLOPs reduction.

\setlength{\tabcolsep}{3pt}
\begin{table}
	\begin{center}
		\begin{tabular}{l||cccc}
			\hline
		      ResNet-50  & \# Params. $\downarrow$ &  \# FLOPs $\downarrow$ & Top-1  & Top-5 \\
		      \hline\hline
		      GCP-f & $1.12 \times$ & $1.55 \times$ & $-0.78$ & $-0.31$ \\
			\hline \hline      
		      GCP-f & $1.32 \times$ & $2.24 \times$& $+0.16$  & $+0.03$ \\
			\hline \hline      
		      GCP-f & $2.18 \times$ & $3.47 \times$& $+0.98$  & $+0.23$ \\		      
			\hline \hline      
		      GCP-f & $3.09 \times$ & $5 \times$& $+2.18$  & $+0.67$ \\	
			\hline \hline      
		      GCP-f & $5.75 \times$ & $10 \times$& $+7.53$  & $+3.87$ \\			      
		\end{tabular}
	\end{center}
	\vspace{-1.5em}
\caption{Comparisons on Caltech-$256$ for ResNet-$50$. The top-$1$ and top-$5$ error \% of the pre-trained model are $18.96$ and $9.73$ respectively.}
\label{tab:caltech256}
\end{table}

\section{Discussion}
We note that our choice of DCP \cite{zhuang2018discrimination} as the main baseline is due to the fact that they achieve the state-of-the-art results of compressing ResNet50 on ImageNet via channel pruning, and not due to our method being an alternative to theirs. We in fact view our method as complementary the theirs in that it provides a clean way to learn the required width of each layer. It is in fact reasonable to assume that the specialized manner in which they optimize the weights of their network is advantageous compared to ours and that accounts for their superiority in the regime of mild pruning. However, once we require a more aggressive pruning, our method is superior due to increased gain from non-uniform pruning. An interesting follow up would be a method combining the techniques, using our methods for attaining the architecture, combined with alternative methods for tuning the weights of the new network.

{\small
\bibliographystyle{plain}
\bibliography{main}

\begin{thebibliography}{10}

\bibitem{alvarez2016learning}
Jose~M Alvarez and Mathieu Salzmann.
\newblock Learning the number of neurons in deep networks.
\newblock In {\em Advances in Neural Information Processing Systems}, pages
  2270--2278, 2016.

\bibitem{carreira2018learning}
Miguel~A Carreira-Perpin{\'a}n and Yerlan Idelbayev.
\newblock “learning-compression” algorithms for neural net pruning.
\newblock In {\em Proceedings of the IEEE Conference on Computer Vision and
  Pattern Recognition}, pages 8532--8541, 2018.

\bibitem{courbariaux2015binaryconnect}
Matthieu Courbariaux, Yoshua Bengio, and Jean-Pierre David.
\newblock Binaryconnect: Training deep neural networks with binary weights
  during propagations.
\newblock In {\em Advances in neural information processing systems}, pages
  3123--3131, 2015.

\bibitem{deng2009imagenet}
Jia Deng, Wei Dong, Richard Socher, Li-Jia Li, Kai Li, and Li~Fei-Fei.
\newblock Imagenet: A large-scale hierarchical image database.
\newblock In {\em Computer Vision and Pattern Recognition, 2009. CVPR 2009.
  IEEE Conference on}, pages 248--255. Ieee, 2009.

\bibitem{girshick2014rich}
Ross Girshick, Jeff Donahue, Trevor Darrell, and Jitendra Malik.
\newblock Rich feature hierarchies for accurate object detection and semantic
  segmentation.
\newblock In {\em Proceedings of the IEEE conference on computer vision and
  pattern recognition}, pages 580--587, 2014.

\bibitem{gong2014compressing}
Yunchao Gong, Liu Liu, Ming Yang, and Lubomir Bourdev.
\newblock Compressing deep convolutional networks using vector quantization.
\newblock {\em arXiv preprint arXiv:1412.6115}, 2014.

\bibitem{han2015deep}
Song Han, Huizi Mao, and William~J Dally.
\newblock Deep compression: Compressing deep neural networks with pruning,
  trained quantization and huffman coding.
\newblock {\em arXiv preprint arXiv:1510.00149}, 2015.

\bibitem{han2015learning}
Song Han, Jeff Pool, John Tran, and William Dally.
\newblock Learning both weights and connections for efficient neural network.
\newblock In {\em Advances in neural information processing systems}, pages
  1135--1143, 2015.

\bibitem{hanson1989comparing}
Stephen~Jos{\'e} Hanson and Lorien~Y Pratt.
\newblock Comparing biases for minimal network construction with
  back-propagation.
\newblock In {\em Advances in neural information processing systems}, pages
  177--185, 1989.

\bibitem{hassibi1993second}
Babak Hassibi and David~G Stork.
\newblock Second order derivatives for network pruning: Optimal brain surgeon.
\newblock In {\em Advances in neural information processing systems}, pages
  164--171, 1993.

\bibitem{he2016deep}
Kaiming He, Xiangyu Zhang, Shaoqing Ren, and Jian Sun.
\newblock Deep residual learning for image recognition.
\newblock In {\em Proceedings of the IEEE conference on computer vision and
  pattern recognition}, pages 770--778, 2016.

\bibitem{he2017channel}
Yihui He, Xiangyu Zhang, and Jian Sun.
\newblock Channel pruning for accelerating very deep neural networks.
\newblock In {\em International Conference on Computer Vision (ICCV)},
  volume~2, 2017.

\bibitem{HofferBGS18}
Elad Hoffer, Ron Banner, Itay Golan, and Daniel Soudry.
\newblock Norm matters: efficient and accurate normalization schemes in deep
  networks.
\newblock In {\em NeurIPS}, pages 2164--2174, 2018.

\bibitem{hu2016network}
Hengyuan Hu, Rui Peng, Yu-Wing Tai, and Chi-Keung Tang.
\newblock Network trimming: A data-driven neuron pruning approach towards
  efficient deep architectures.
\newblock {\em arXiv preprint arXiv:1607.03250}, 2016.

\bibitem{hubara2017quantized}
Itay Hubara, Matthieu Courbariaux, Daniel Soudry, Ran El-Yaniv, and Yoshua
  Bengio.
\newblock Quantized neural networks: Training neural networks with low
  precision weights and activations.
\newblock {\em Journal of Machine Learning Research}, 18(187):1--30, 2017.

\bibitem{jaderberg2014speeding}
Max Jaderberg, Andrea Vedaldi, and Andrew Zisserman.
\newblock Speeding up convolutional neural networks with low rank expansions.
\newblock {\em arXiv preprint arXiv:1405.3866}, 2014.

\bibitem{khetan2020schubert}
Ashish Khetan and Zohar Karnin.
\newblock schubert: Optimizing elements of bert.
\newblock {\em arXiv preprint arXiv:2005.06628}, 2020.

\bibitem{krizhevsky2012imagenet}
Alex Krizhevsky, Ilya Sutskever, and Geoffrey~E Hinton.
\newblock Imagenet classification with deep convolutional neural networks.
\newblock In {\em Advances in neural information processing systems}, pages
  1097--1105, 2012.

\bibitem{lan2019albert}
Zhenzhong Lan, Mingda Chen, Sebastian Goodman, Kevin Gimpel, Piyush Sharma, and
  Radu Soricut.
\newblock Albert: A lite bert for self-supervised learning of language
  representations.
\newblock {\em arXiv preprint arXiv:1909.11942}, 2019.

\bibitem{lebedev2014speeding}
Vadim Lebedev, Yaroslav Ganin, Maksim Rakhuba, Ivan Oseledets, and Victor
  Lempitsky.
\newblock Speeding-up convolutional neural networks using fine-tuned
  cp-decomposition.
\newblock {\em arXiv preprint arXiv:1412.6553}, 2014.

\bibitem{li2016pruning}
Hao Li, Asim Kadav, Igor Durdanovic, Hanan Samet, and Hans~Peter Graf.
\newblock Pruning filters for efficient convnets.
\newblock {\em arXiv preprint arXiv:1608.08710}, 2016.

\bibitem{liu2018darts}
Hanxiao Liu, Karen Simonyan, and Yiming Yang.
\newblock Darts: Differentiable architecture search.
\newblock {\em arXiv preprint arXiv:1806.09055}, 2018.

\bibitem{liu2017learning}
Zhuang Liu, Jianguo Li, Zhiqiang Shen, Gao Huang, Shoumeng Yan, and Changshui
  Zhang.
\newblock Learning efficient convolutional networks through network slimming.
\newblock In {\em Computer Vision (ICCV), 2017 IEEE International Conference
  on}, pages 2755--2763. IEEE, 2017.

\bibitem{long2015fully}
Jonathan Long, Evan Shelhamer, and Trevor Darrell.
\newblock Fully convolutional networks for semantic segmentation.
\newblock In {\em Proceedings of the IEEE conference on computer vision and
  pattern recognition}, pages 3431--3440, 2015.

\bibitem{luo2017thinet}
Jian-Hao Luo, Jianxin Wu, and Weiyao Lin.
\newblock Thinet: A filter level pruning method for deep neural network
  compression.
\newblock {\em arXiv preprint arXiv:1707.06342}, 2017.

\bibitem{parikh2014proximal}
Neal Parikh, Stephen Boyd, et~al.
\newblock Proximal algorithms.
\newblock {\em Foundations and Trends{\textregistered} in Optimization},
  1(3):127--239, 2014.

\bibitem{rastegari2016xnor}
Mohammad Rastegari, Vicente Ordonez, Joseph Redmon, and Ali Farhadi.
\newblock Xnor-net: Imagenet classification using binary convolutional neural
  networks.
\newblock In {\em European Conference on Computer Vision}, pages 525--542.
  Springer, 2016.

\bibitem{simonyan2014two}
Karen Simonyan and Andrew Zisserman.
\newblock Two-stream convolutional networks for action recognition in videos.
\newblock In {\em Advances in neural information processing systems}, pages
  568--576, 2014.

\bibitem{singh2019darc}
Shashank Singh, Ashish Khetan, and Zohar Karnin.
\newblock Darc: Differentiable architecture compression.
\newblock {\em arXiv preprint arXiv:1905.08170}, 2019.

\bibitem{srinivas2017training}
Suraj Srinivas, Akshayvarun Subramanya, and R~Venkatesh Babu.
\newblock Training sparse neural networks.
\newblock In {\em 2017 IEEE Conference on Computer Vision and Pattern
  Recognition Workshops (CVPRW)}, pages 455--462. IEEE, 2017.

\bibitem{tai2015convolutional}
Cheng Tai, Tong Xiao, Yi~Zhang, Xiaogang Wang, et~al.
\newblock Convolutional neural networks with low-rank regularization.
\newblock {\em arXiv preprint arXiv:1511.06067}, 2015.

\bibitem{vinyals2015show}
Oriol Vinyals, Alexander Toshev, Samy Bengio, and Dumitru Erhan.
\newblock Show and tell: A neural image caption generator.
\newblock In {\em Proceedings of the IEEE conference on computer vision and
  pattern recognition}, pages 3156--3164, 2015.

\bibitem{wen2016learning}
Wei Wen, Chunpeng Wu, Yandan Wang, Yiran Chen, and Hai Li.
\newblock Learning structured sparsity in deep neural networks.
\newblock In {\em Advances in Neural Information Processing Systems}, pages
  2074--2082, 2016.

\bibitem{zhu2016trained}
Chenzhuo Zhu, Song Han, Huizi Mao, and William~J Dally.
\newblock Trained ternary quantization.
\newblock {\em arXiv preprint arXiv:1612.01064}, 2016.

\bibitem{zhuang2018discrimination}
Zhuangwei Zhuang, Mingkui Tan, Bohan Zhuang, Jing Liu, Yong Guo, Qingyao Wu,
  Junzhou Huang, and Jinhui Zhu.
\newblock Discrimination-aware channel pruning for deep neural networks.
\newblock In {\em Advances in Neural Information Processing Systems}, pages
  883--894, 2018.

\end{thebibliography}
}

\end{document}